\pdfoutput=1
\documentclass{nomospreprint}
\makeatletter
\@ifundefined{if@biographyTOCentrynotmade}{\newif\if@biographyTOCentrynotmade\@biographyTOCentrynotmadetrue}{}
\@ifundefined{c@biography}{}{}
\makeatother
\ifdefined\pdfpagewidth\fi
\makeatletter
\def\ps@headings{\let\@oddhead\@empty\let\@evenhead\@empty\def\@oddfoot{\hfil{\footerpagefont\thepage}\hfil}\let\@evenfoot\@oddfoot}
\let\ps@titlepage\ps@headings
\makeatother
\makeatletter
\def\restoreSection{%
\def\section{\@startsection{section}{1}{\z@}{-2.0ex \@plus -0.5ex \@minus -0.2ex}{1.5ex \@plus 0.3ex \@minus 0.2ex}{\normalfont\large\bfseries\raggedright}}%
\def\subsection{\@startsection{subsection}{2}{\z@}{-1.8ex \@plus -0.5ex \@minus -0.2ex}{0.8ex \@plus 0.2ex}{\normalfont\normalsize\bfseries\raggedright}}%
\def\subsubsection{\@startsection{subsubsection}{3}{\z@}{-1.5ex \@plus -0.5ex \@minus -0.2ex}{0.5ex \@plus 0.2ex}{\normalfont\normalsize\bfseries\raggedright}}}
\restoreSection
\def\thesection{\arabic{section}}

\def\@seccntformat#1{\csname the#1\endcsname\quad}
\def\refname{References}
\makeatother
\makeatletter
\def\tablename{Table}
\def\figurename{Figure}
\def\nomoscapfont{\normalfont\small}
\long\def\nomos@caption#1#2{%
\sbox\@tempboxa{\nomoscapfont #1: #2}%
\ifdim\wd\@tempboxa>\hsize{\nomoscapfont #1: #2\par}%
\else\global\@minipagefalse\hb@xt@\hsize{\hfil\box\@tempboxa\hfil}\fi}
\long\def\@makecaption#1#2{%
\ifx\@captype\@IEEEtablestring\vskip 2pt\nomos@caption{#1}{#2}\vskip 6pt
\else\vskip 8pt\nomos@caption{#1}{#2}\vskip 2pt\fi}
\makeatother
\makeatletter\def\@textbottom{\vskip \z@ \@plus 12pt}\makeatother
\newenvironment{nomoslist}{\begin{list}{}{\setlength{\leftmargin}{2em}\setlength{\labelwidth}{1.6em}\setlength{\labelsep}{0.4em}\setlength{\itemsep}{0pt}\setlength{\topsep}{2pt}\setlength{\parsep}{4pt}}}{\end{list}}
\def\bodyfont{\rmfamily\fontencoding{T1}\fontseries{m}\fontsize{10}{12}\selectfont}
\def\referencefont{\rmfamily\fontencoding{T1}\fontseries{m}\fontsize{7.61}{9}\selectfont}

\usepackage{amsmath,amssymb}
\usepackage{array}
\usepackage{booktabs}
\usepackage{graphicx}
\usepackage{url}
\usepackage{multirow}
\usepackage{cite}
\usepackage[capitalize,noabbrev]{cleveref}

\def\xfigwd{0pt}

\makeatletter
\def\appendices{\par%
    \def\theHsection{Appendix.\Alph{section}}%
    \xdef\Hy@chapapp{appendix}%
    \setcounter{section}{-1}%
    \setcounter{subsection}{0}%
    \setcounter{subsubsection}{0}%
    \setcounter{paragraph}{0}%
    \def\thesection{\Alph{section}}%
    \def\thesectiondis{APPENDIX \Alph{section}}%
    \refstepcounter{section}%
    \setcounter{section}{0}%
}
\makeatother

\begin{document}

\twocolumn[{%
\begin{center}
\rule{\textwidth}{4pt}\par
\vspace{0.22in}
{\fontsize{17.28}{20.5}\selectfont\rmfamily\bfseries Enhancing sLLM Translation for the Construction Domain via Fine-tuning and Retrieval-Augmented Generation\par}
\vspace{0.24in}
\rule{\textwidth}{1pt}\par
\vspace{0.3in}
{\rmfamily\fontsize{10}{12}\selectfont \begin{tabular}[t]{c}\bfseries Min-Young Yu\\ Dept.\ of Applied Artificial Intelligence\\ Seoul National University of Science and Technology\\ \texttt{minyoung@seoultech.ac.kr}\end{tabular}\hspace{0.45in}\begin{tabular}[t]{c}\bfseries Hanul Kim\\ Dept.\ of Applied Artificial Intelligence\\ Seoul National University of Science and Technology\\ \texttt{hukim@seoultech.ac.kr}\end{tabular}\par}
\vspace{0.32in}
{\rmfamily\fontsize{12}{14}\selectfont\bfseries Abstract\par}
\vspace{1.6ex}
\begin{minipage}{0.72\textwidth}
\rmfamily\fontsize{10}{12}\selectfont
Communication barriers between site managers and a growing foreign workforce are a serious safety risk on construction sites.
General-purpose translation applications misread construction-specific terminology, including Japanese-derived site slang, and such errors can become a latent cause of serious accidents.
We propose a real-time, domain-specific translation system that combines a small language model (sLLM) with retrieval-augmented generation (RAG).
Using LLM-based synthetic data generation, we build a Korean--English parallel corpus of 750 sentences centered on 50 key construction terms, and compare six configurations: a multilingual NMT baseline (M2M100 1.2B), its LoRA fine-tuned variant, a general-purpose sLLM (Gemma 3 4B), a fine-tuned Gemma 3 4B, the fine-tuned model with RAG, and the proposed Gemma 3 4B with RAG, which retrieves exact term translations from an external knowledge base.
On 250 held-out sentences, the proposed model attains the highest semantic accuracy (COMET 0.846, against 0.833 for the fine-tuned sLLM and 0.770 for the zero-shot sLLM) without any parameter update, while fine-tuning yields the highest lexical overlap (BLEU 23.35, chrF 59.20).
Adding RAG to an already fine-tuned model brings no gain, which we interpret as redundant knowledge.
A qualitative case study of a safety-relevant term shows that RAG avoids the transliteration and unrelated-word errors of the other systems.
These results suggest that RAG is an efficient and easily maintainable route to on-device, domain-specific translation in safety-critical settings.
\end{minipage}
\end{center}
\vspace{0.28in}
}]
\thispagestyle{empty}
{\let\thefootnote\relax\footnotetext{This paper is an English adaptation of the first author's M.Eng.\ thesis (Seoul National University of Science and Technology, February 2026).}}
{\let\thefootnote\relax\footnotetext{Preprint.}}

\section{Introduction}

\IEEEPARstart{T}{he} Korean construction industry is undergoing an unprecedented demographic transition.
A shrinking, ageing domestic workforce and a growing aversion to so-called ``3D'' (dirty, dangerous, difficult) jobs have produced a chronic labor shortage, which is being filled rapidly by foreign workers.
This is more than a substitution of labor: it changes the basic operating structure of a construction site.
According to the Construction Workers Mutual Aid Association \cite{kcwa}, about 229,500 foreign workers, 14.7\% of the workforce, were counted on construction sites in 2024, up from 11.8\% in 2020.
The same survey indicates that many of them are ethnic Koreans from China, so a Korean-language barrier does not apply to every foreign worker; but a growing number of workers speak Vietnamese, Thai, Khmer, Russian, or other languages, which makes the barrier a rising concern.

This shift introduces a new hazard: the language barrier.
A construction site mixes heavy equipment, work at height, and hazardous chemicals, so immediate and exact communication of work orders, safety warnings, and emergency alerts is a matter of life and death.
Linguistic misunderstanding disables these safety mechanisms.
We view this as a structural problem, the collapse of the channel that recognizes and conveys hazards, rather than a matter of individual carelessness or ignorance of safety rules.

\textbf{Limits of existing solutions.}
Communication tools in current use fall short of this safety problem in three ways.
(i)~\emph{Errors of general-purpose NMT.}
Translation applications such as Google Translate and Naver Papago are trained on vast data but are weak on domain terminology.
They transliterate the term for squared timber used as a brace as ``gakumok'' instead of translating it, or render \emph{gombang} (carrying materials by hand) as ``bear room.''
Where safety is at stake, such errors are not mere failures of information transfer but potential causes of accidents.
(ii)~\emph{Network dependence and security.}
Translation with high-performance LLMs (e.g., GPT-4) relies on cloud APIs and therefore on an Internet connection, which is unavailable in underground tunnels, inside high-rise buildings, and at remote sites.
Sending site drawings or confidential documents to an external server is also a serious weakness for corporate security.
(iii)~\emph{Inefficiency of human interpretation.}
Stationing professional interpreters is costly, and covering Vietnamese, Thai, Khmer, Russian, and other languages in real time on a multinational site is physically impossible.

\textbf{This work.}
We propose an sLLM-based RAG translation system that offers both the accuracy needed to ensure \emph{safety} and the availability needed for immediate use \emph{on site}.
Built on Google's Gemma 3 4B, it is designed to run in real time on handheld devices.
Our contributions are as follows.
\begin{nomoslist}
\item[(1)] \textbf{Data construction methodology.} For the low-resource construction domain, we design a pipeline that builds a high-quality synthetic parallel corpus with LLMs. We construct 750 Korean--English sentences around 50 key site terms (\cref{sec:data}).
\item[(2)] \textbf{Edge-oriented model validation.} Using the recent lightweight Gemma 3 4B, we examine whether on-device AI that runs on a mobile device without a large server is effective for this task (\cref{sec:method}).
\item[(3)] \textbf{Comparison of adaptation techniques.} We compare the two mainstream approaches to domain adaptation, parametric adaptation (fine-tuning, LoRA) and non-parametric knowledge injection (RAG), with COMET, BLEU, and chrF and with a qualitative error analysis (\cref{sec:results}).
\end{nomoslist}
\par
The remainder of this paper is organized as follows.
\cref{sec:related} reviews related work, \cref{sec:prelim} provides background on site terminology, parameter-efficient fine-tuning, retrieval-augmented generation, and evaluation metrics, \cref{sec:method} describes the proposed system, the dataset, and the compared models, \cref{sec:results} presents quantitative and qualitative results, and \cref{sec:conclusion} discusses limitations and future work.

\section{Related Work}
\label{sec:related}

\subsection{From Statistical to Neural Machine Translation}

Machine translation has progressed from rule-based systems through statistical machine translation (SMT) to neural machine translation (NMT), and each transition changed how language is modeled mathematically.
Phrase-based SMT \cite{koehn2003} splits a sentence into phrases, scores their alignments, and combines a translation model with a language model through Bayes' rule, which sharply improved accuracy within local contexts.
Because phrases are treated as independent units, however, such models ignore sentence-level context \cite{chiang2005}.
For language pairs with very different word order, such as Korean and English, or for long-distance dependencies, they reorder by statistical frequency rather than linguistic structure and yield unnatural or fragmented output.
SMT also suffers from data sparsity: n-grams absent from the training data receive zero or heavily smoothed probabilities.
In a specialized domain such as construction, where terminology and slang never seen in general corpora are frequent, this appears as the out-of-vocabulary (OOV) problem, and subword units were later proposed to address rare words \cite{sennrich2016}.

Sequence-to-sequence learning \cite{sutskever2014} replaced this pipeline with an end-to-end network in which an encoder (an RNN, LSTM, or GRU) compresses a variable-length input into a fixed-size context vector and a decoder generates the target sequence.
Quality of these early models degrades as sentences grow longer; the fixed-size vector acts as an information bottleneck, and vanishing gradients hamper the learning of long-term dependencies \cite{cho2014properties}.
The Transformer \cite{vaswani2017} removes recurrence and convolution entirely and relies on attention alone.
Self-attention relates every token to every other token through matrix operations, so the path length between any two positions is constant, $O(1)$, instead of $O(n)$ in an RNN, and all tokens can be processed in parallel.
Prepending a target-language token lets a single model learn several language pairs at once \cite{johnson2017}; knowledge learned from high-resource pairs then transfers to low-resource pairs, an observation that matters for a data-poor domain such as construction.
In the authors' human side-by-side evaluation on simple sentence sets, Google's GNMT reduced translation errors by 60\% on average relative to Google's production phrase-based system \cite{wu2016}.

\subsection{Machine Translation Adaptation for Specialized Domains}

\textbf{Model-centric adaptation.}
Training from scratch on little data is inefficient, so the standard strategy is to initialize from a model pre-trained on large general corpora and continue training on a small in-domain set.
Luong and Manning \cite{luong2015} showed that a few epochs of in-domain training sharply raise BLEU.
Over-optimizing on domain data, however, risks \emph{catastrophic forgetting} \cite{french1999}: fluency drops, and quality on out-of-domain input collapses.
Remedies include mixed fine-tuning of domain and general data \cite{chu2017} and regularization that keeps parameters close to their original values, such as elastic weight consolidation \cite{kirkpatrick2017}.

\textbf{Constrained translation and terminology consistency.}
In domains tied to safety, exact terminology often matters more than fluency, yet a probabilistic generator may drop a term or replace it with a synonym.
Dynamic beam allocation \cite{post2018} forces user-specified words or phrases into the output during beam search; it guarantees consistency by pruning hypotheses that lack the terms, but it is a hard constraint applied at decoding time.
Soft constraints inject target terms into the training data, for example by appending them to the source sentence as inline annotations, so that the model learns to copy them \cite{dinu2019}.
This does not raise run-time complexity and is attractive for real-time industrial translation.

\subsection{LLM-based Translation}

Large language models (LLMs) can translate without parameter updates.
With in-context learning, a handful of source--target examples placed in the prompt are enough for the model to infer the task; Brown et al.\ \cite{brown2020} reported that few-shot translation quality improves as the number of demonstrations grows, which is attractive where there are too few data to train a supervised system.
Because LLMs accept long context windows, they can also use document-level context, which improves consistency over sentence-level translation, although critical errors persist \cite{karpinska2023}.
Trained on multilingual tokens, LLMs further acquire cross-lingual transfer: knowledge learned in one language carries over to tasks in another \cite{lin2022}.
Evaluations of GPT models show that they are competitive with commercial systems on high-resource language pairs but lag on low-resource and distant pairs \cite{jiao2023,hendy2023}, so general ability alone does not remove the need for domain support.
Instruction tuning \cite{wei2022} allows control of register and style, for instance by asking for polite, colloquial, or report-style output.
In specialized domains, where ordinary words take unusual meanings (``nut'' as a fastener rather than food), supplying domain information in the prompt before translation, such as dictionary hints for the words in the source sentence or retrieved examples, can improve quality without additional training \cite{ghazvininejad2023,moslem2023}.

\subsection{Construction-Domain Translation}

Most MT research targets formal text such as news and meeting minutes; the language of construction sites and similar industrial settings has long been a blind spot because of data scarcity and linguistic difficulty.
Four properties set it apart.
First, it behaves like a closed sociolect: it contains many Japanese-derived words and site-specific slang and abbreviations (\cref{sec:jp}), such as \emph{gonguri} (concrete), \emph{narasi} (leveling), and \emph{bansaeng} (binding wire), often in preference to standard Korean words.
General-purpose models trained on standard-language corpora treat such words as out-of-vocabulary (OOV) and may mistake them for proper nouns or render them as unrelated words.
Second, written registers such as specifications and safety manuals coexist with fragmented spoken registers (work orders, radio calls) in which subjects and objects are dropped and word order is ungrammatical; the intent of an utterance (command, request, or warning) is then left implicit.
Third, translation in a safety-critical setting demands more than average quality: meaning reversals, such as turning ``do not open the valve'' into ``open the valve,'' must be driven toward zero, so semantic preservation matters more than surface overlap.
Fourth, commercial translators trained on general web data suffer from domain mismatch \cite{koehn2017}: a model prefers the most frequent sense, so ``nut'' (a fastener) may become a food and ``beam'' (a structural member) a light ray.
Earlier domain-adaptation attempts do not fully resolve this: glossary enforcement in the SMT era tended to hurt fluency, and early NMT fine-tuning on small in-domain data risked overfitting and forgetting.
A survey of domain adaptation \cite{saunders2022} discusses how little parallel data exist in many specialized domains and the limits of plain fine-tuning there, which motivates architectures that inject external knowledge, such as LLMs with RAG.

\section{Preliminaries}
\label{sec:prelim}

\subsection{Japanese-mediated Site Terminology}
\label{sec:jp}

Many terms used on Korean construction sites are Japanese-derived, and public bodies have run campaigns to replace them with Korean words; the National Institute of Korean Language compiled a collection of such terms for purification \cite{nikl2005}, and a 2019 agreement between the Korea Land and Housing Corporation and the Institute launched a campaign to rewrite site terms in plain Korean \cite{lh2019}.
Such words were reshaped by the Korean sound system.
\emph{Gonguri} (concrete), for example, can be traced to English \emph{concrete}, which became \emph{konkuriito} in Japanese and was simplified to \emph{gonguri} on sites instead of settling as the standard loanword \emph{konkeuriteu}.
\emph{Yarikkiri} (a piece-rate arrangement in which workers leave once the assigned work is done) combines the Japanese verbs \emph{yaru} ``to do'' and \emph{kiru} ``to finish,'' and \emph{narasi} (leveling the ground) comes from \emph{narasu} ``to level'' without any change in sound.
Phonologically they show typical traits of site slang: shortening of multisyllabic loanwords and tensification of consonants.
Because such words are passed on by ear rather than in text, they are poorly covered by the textual corpora used to train translation models.

\subsection{Transformer and Attention}

The Transformer's core operation, scaled dot-product attention, takes query ($Q$), key ($K$), and value ($V$) matrices and computes
\begin{equation}
\mathrm{Attention}(Q,K,V)=\mathrm{softmax}\!\left(\frac{QK^{\top}}{\sqrt{d_k}}\right)V ,
\label{eq:attn}
\end{equation}
where $d_k$ is the key dimension.
The $1/\sqrt{d_k}$ scaling is necessary for stable training: when $d_k$ is large, the dot products grow in magnitude and push the softmax into regions with near-zero gradient \cite{vaswani2017}.
Multi-head attention projects $Q$, $K$, and $V$ with $h$ different linear maps into separate subspaces and attends in parallel, letting the model look at different positions from several viewpoints.
Because the architecture has no recurrence, positional information is injected through sinusoidal positional encodings added to the input embeddings.

\subsection{Parameter-Efficient Fine-tuning}

The pre-train/fine-tune paradigm \cite{devlin2019} internalizes grammatical and semantic knowledge during pre-training and then fits the distribution of task-specific data.
The theoretical basis for parameter-efficient fine-tuning (PEFT) is the \emph{intrinsic dimension} hypothesis: the larger a pre-trained model, the lower the intrinsic dimension of the objective needed to solve a task, so a low-dimensional projection can reach more than 90\% of full fine-tuning performance \cite{aghajanyan2021}.
LoRA \cite{hu2022} engineers this idea by freezing the pre-trained weights $W_0\in\mathbb{R}^{d\times k}$ and approximating the update $\Delta W$ by a product of two low-rank matrices $B\in\mathbb{R}^{d\times r}$ and $A\in\mathbb{R}^{r\times k}$ with $r\ll\min(d,k)$:
\begin{equation}
h = W_0x+\Delta Wx = W_0x+\frac{\alpha}{r}BAx ,
\label{eq:lora}
\end{equation}
where $A$ is initialized from a Gaussian and $B$ at zero so that $\Delta W=0$ at the start of training, and $\alpha$ is a scaling constant.
The number of trainable parameters falls sharply (up to a two-thirds reduction in GPU memory for GPT-3 175B), and since $W_0+\frac{\alpha}{r}BA$ can be merged beforehand, inference incurs no extra latency.
QLoRA \cite{dettmers2023} adds 4-bit NormalFloat quantization of the frozen weights, double quantization of the quantization constants (about 0.37 bit per parameter saved), and paged optimizers, and fine-tunes a 65B-parameter model on a single 48\,GB GPU.

\subsection{Retrieval-Augmented Generation}

RAG \cite{lewis2020} combines knowledge stored in model weights (\emph{parametric memory}) with an external corpus accessed through a vector index (\emph{non-parametric memory}).
Given an input $x$, a retriever returns relevant documents $z$, which are fed with $x$ to a generator that produces $y$.
In the RAG-Sequence formulation a single retrieved document is assumed to condition the whole output, and the probability is approximated by summing over the top-$K$ documents,
\begin{equation}
p(y\mid x)\approx\sum_{z\in\mathrm{top\text{-}}K\,p_\eta(\cdot\mid x)} p_\eta(z\mid x)\,p_\theta(y\mid x,z) ,
\label{eq:rag}
\end{equation}
whereas RAG-Token lets a different document influence each token and marginalizes per token.
Retrieval usually relies on dense passage retrieval (DPR) \cite{karpukhin2020}, in which independent BERT encoders map a question and a passage to $d$-dimensional vectors and similarity is the dot product $\mathrm{sim}(q,p)=E_Q(q)^{\top}E_P(p)$.
Compared with sparse retrieval such as TF-IDF and BM25, dense retrieval captures synonyms and paraphrases.
Retrieved text can conflict with the model's parametric knowledge, so reranking with a cross-encoder, natural-language-inference filtering, and context compression are used to reduce hallucination \cite{gao2023}.

\subsection{Evaluation Metrics}

BLEU \cite{papineni2002} and ROUGE score n-gram overlap with a reference, so they penalize a correct translation that uses different words or a different order (lexical mismatch) and correlate poorly with human judgment \cite{zhang2020}.
chrF is a character n-gram F-score \cite{popovic2015}.
BERTScore \cite{zhang2020} compares contextual BERT embeddings of every token pair by cosine similarity with greedy matching and is robust to synonyms and reordering.
COMET \cite{rei2020} goes further: it is trained directly on human direct-assessment scores and takes the source, the hypothesis, and the reference as input, using a pre-trained multilingual encoder such as XLM-RoBERTa, pooling, and concatenation to predict quality.
Because it also sees the source, COMET judges better than earlier metrics whether the meaning of the source is preserved, which is why we treat it as the primary metric for safety-oriented translation.

\section{Proposed System}
\label{sec:method}

\subsection{Overview}

\begin{figure*}[t]
\centering
\includegraphics[width=0.92\textwidth]{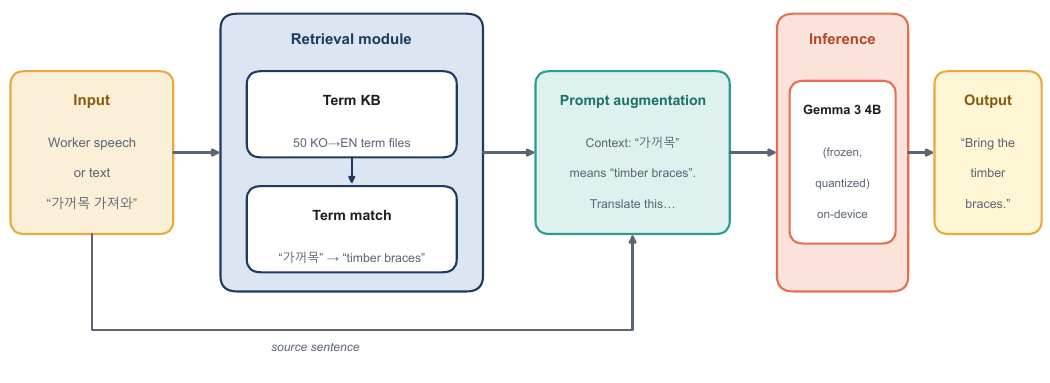}
\caption{Prompt augmentation through RAG. A spoken or typed instruction (here, ``bring the \emph{gakkumok}'') is checked against the term knowledge base; the matched term translation is inserted into the prompt together with the source sentence, and a frozen, quantized Gemma 3 4B model on the device produces the translation.}
\label{fig:rag}
\end{figure*}

The proposed system uses Google's Gemma 3 4B \cite{gemma3} as its translation engine and is lightweight enough to run in real time.
A RAG module is attached to maximize the accuracy of construction terms; its operation is shown in \cref{fig:rag}.
First, we build an external knowledge base (KB) in the form of a simple dictionary of Korean:English pairs for the 50 key terms.
When a translation request arrives, the system searches the input sentence for terms registered in the KB.
If a term is found, the exact translation pair is retrieved and injected into the prompt of the Gemma 3 4B model as an explicit instruction, for example ``Translate the following sentence. Note that \emph{bigye} must be translated as `scaffolding.' '' to guide generation.

This architecture raises translation accuracy while remaining easy to extend.
When a new safety regulation or term is added in construction, fine-tuning (FT) requires parameter re-training, but with RAG it is enough to update a knowledge base kept as text files, and the change takes effect immediately.
This helps respond to fast-changing industrial sites and keeps maintenance cost low.

\subsection{Dataset Construction}
\label{sec:data}

\begin{figure*}[t]
\centering
\includegraphics[width=0.92\textwidth]{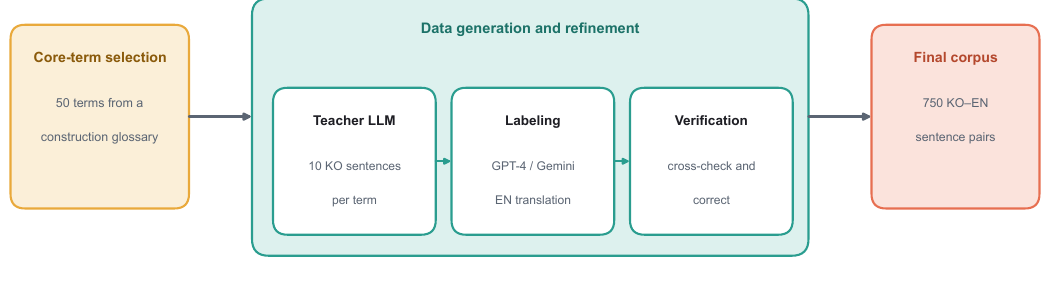}
\caption{Corpus construction process. Fifty core terms are selected; a teacher LLM writes Korean sentences for each term, commercial LLMs provide English translations, and the pairs are cross-checked and refined into the final 750-pair corpus.}
\label{fig:corpus}
\end{figure*}

To build a specialized-domain translation model we applied LLM-based synthetic data generation and augmentation (\cref{fig:corpus}).

\textbf{Key-term selection.}
Referring to reliable sources such as the Korean Building Owners Association's glossary of plain-Korean construction terms, we selected 50 key terms that are frequent on site and prone to mistranslation (\cref{tab:terms}).

\textbf{Source corpus generation.}
To compensate for the absence of a construction parallel corpus, we used GPT-4 as a \emph{teacher model}.
For each of the 50 terms we prompted the strong teacher LLM to write 10 Korean example sentences reflecting realistic contexts such as work orders, safety warnings, and situation reports, giving 750 high-quality Korean sentences in total.

\textbf{Reference translations.}
English translations were obtained for 500 of the generated sentences with two top commercial LLMs, ChatGPT (GPT-4 based) and Gemini Pro.
The outputs were cross-reviewed and corrected where necessary, and the most accurate and natural translation was fixed as the ground truth, which gives the effect of knowledge distillation.

Gunasekar et al.\ \cite{gunasekar2023} showed that data quality matters more than quantity: synthetic, textbook-quality data generated by an LLM improves model performance more than noisy web data.
We expect the same in construction.
Rather than training on the raw language of the site, it is more efficient to generate a refined and structured ``construction-terminology textbook'' with an LLM and train on it.

\subsection{Term-level Knowledge Decomposition for RAG}

To interpret domain-specific terms (site slang and technical terms) precisely and improve multilingual translation, we manage knowledge as one Markdown file per term (\emph{term-level knowledge decomposition}).
Each file is named after the source term (e.g., \texttt{darukki.md} for \emph{darukki}) and holds the following structured fields: the original term, the standard Korean counterpart, the English translation (standard technical term), example sentences from site dialogue and technical documents, and optionally similar or confusable terms.

This file-per-term layout serves three purposes.
(i)~\emph{Minimal retrieval unit.} Separating knowledge by term rather than by document removes unnecessary context noise from the RAG pipeline and improves retrieval accuracy.
(ii)~\emph{Translation-oriented knowledge injection.} Supplying the sense-based translation actually used in construction, rather than a plain dictionary gloss, mitigates the semantic distortion of general translation models.
(iii)~\emph{Dynamic semantic lookup.} When a term is detected in the user's input, the Markdown file with the same name is designated as a priority retrieval target and injected into the context.

The RAG flow is: (1) extract term candidates from the input sentence by morphological analysis or NER; (2) match them against a pre-built term index; (3) retrieve the matching \texttt{\{term\}.md} first; (4) insert that document into the LLM prompt as retrieval context; (5) generate the translation, explanation, or sentence on the basis of that information.
The system thereby gains higher domain-specific translation accuracy, fixed meanings for ambiguous expressions, a shorter RAG context with better token efficiency, and the ability to add a new term by adding a file.

\subsection{Compared Models}
\label{sec:models}

Of the 750 constructed pairs, 500 were used for training (fine-tuning) and the remaining 250 served as a separate test set for an objective evaluation.
To verify the proposed model we compare the six configurations in \cref{tab:models}; M1 and M2 use the multilingual M2M100 model \cite{fan2021}.
In M6, the proposed model, Gemma 3 is kept frozen, key terms in the input are retrieved from the knowledge base, and each is injected into the prompt in the form ``Context: `[term]' means `[translation]'.''

\begin{table*}[t]
\centering
\small
\caption{The six model configurations compared.}
\label{tab:models}
\begin{tabular}{llllp{6.2cm}}
\toprule
ID & Model & Base model & Training & Purpose \\
\midrule
M1 & Baseline & M2M100 1.2B & Zero-shot & Measures the baseline performance of an existing multilingual NMT model \\
M2 & LoRA FT & M2M100 1.2B & LoRA fine-tuning & Measures the effect of parameter adaptation with a small dataset \\
M3 & sLLM baseline & Gemma 3 4B & Zero-shot & Evaluates the built-in translation and reasoning ability of a recent sLLM \\
M4 & sLLM FT & Gemma 3 4B & Full fine-tuning & Gives the upper bound of internalizing domain knowledge in the weights \\
M5 & sLLM FT+RAG & Gemma 3 4B & FT + RAG (zero-shot) & Analyzes knowledge redundancy and conflict \\
M6 & Proposed & Gemma 3 4B & RAG (zero-shot) & Verifies semantic accuracy and efficiency through external knowledge injection \\
\bottomrule
\end{tabular}
\end{table*}

\section{Experiments and Results}
\label{sec:results}

\subsection{Quantitative Results}

The six models were evaluated on the 250-sentence test set with three standard metrics: lexical similarity (BLEU), character-level similarity (chrF), and semantic accuracy (COMET).
Higher is better for all metrics.
Results are given in \cref{tab:results}.

\begin{table*}[t]
\centering
\small
\caption{Translation performance of each model on the 250-sentence test set. Best value per metric in bold.}
\label{tab:results}
\setlength{\tabcolsep}{4pt}
\begin{tabular}{llccc}
\toprule
Model & Parameters & COMET$\uparrow$ & BLEU$\uparrow$ & chrF$\uparrow$ \\
\midrule
M1: M2M100 baseline & 1.2B & 0.6702 & 10.45 & 34.08 \\
M2: M2M100 + LoRA & 1.2B + 4M & 0.6913 & 10.33 & 37.67 \\
M3: Gemma 3 4B (zero-shot) & 4B & 0.7702 & 15.64 & 47.26 \\
M4: Gemma 3 4B FT & 4B (tuned) & 0.8329 & \textbf{23.35} & 59.20 \\
M5: Gemma 3 4B FT + RAG & 4B (tuned) & 0.8323 & 23.23 & \textbf{59.27} \\
M6: Gemma 3 4B + RAG (ours) & 4B & \textbf{0.8462} & 15.66 & 55.90 \\
\bottomrule
\end{tabular}
\end{table*}

\subsection{Analysis}

\textbf{Effect of domain-adaptive training.}
The LoRA fine-tuned M2M100 (M2) improves over the baseline (M1) in COMET (+0.021) and chrF (+3.59), while BLEU is essentially unchanged ($-$0.12).
This suggests that PEFT on a small dataset yields a semantic gain without full retraining.

\textbf{Potential of an sLLM.}
The zero-shot Gemma 3 4B (M3, COMET 0.7702) scores far above both the M2M100 baseline (0.6702) and the LoRA-tuned M2M100 (0.6913).
A recent lightweight language model thus already carries strong general translation ability from its large-scale pre-training and is a sound base model for a domain-specific solution.

\textbf{Fluency from fine-tuning.}
The fine-tuned Gemma 3 4B (M4) reaches the highest BLEU (23.35) among the models without RAG and a chrF of 59.20, close to the best chrF overall.
Fine-tuning internalizes domain knowledge in the weights and is therefore the strongest way to imitate the style and n-gram patterns of the training data; it raises overall fluency.

\textbf{Knowledge redundancy and conflict.}
Adding RAG to the fine-tuned model (M5) brings no gain, and COMET drops slightly (0.8329 to 0.8323).
We interpret this as \emph{knowledge redundancy or conflict}: when the model has already internalized the construction terms through fine-tuning, re-injecting the same information through the prompt may act as noise or disturb the best inference path.
The analysis supports using RAG either as a replacement for expensive fine-tuning or to add factual grounding to a lightweight model whose knowledge has not been internalized.

\textbf{The proposed sLLM+RAG model.}
The proposed Gemma 3 4B + RAG (M6) records the highest COMET of all models (0.8462), 0.076 above the zero-shot model and 0.013 above the fine-tuned one, with no fine-tuning.
M6's BLEU (15.66) is about equal to the zero-shot model's, but its chrF rises from 47.26 to 55.90.
Semantic accuracy is the factor that matters most for construction-site safety, and RAG combines the fluency of Gemma 3 4B with externally verified knowledge in real time.
M6 therefore achieves the best semantic accuracy while remaining the most maintainable and extensible option; it does not, however, match the lexical overlap with the references (BLEU, chrF) of the fine-tuned models.
The differences are measured on a single 250-sentence test set and are not accompanied by significance tests.

\subsection{Qualitative Analysis}

To back the quantitative results and make the reliability gap on specialized terms explicit, we examine the translations of a sentence containing the high-risk term \emph{gakkumok}, \cref{tab:qual}.
The source sentence is ``Jeojjok byeokche sujigi an majeuni gakkumogeul daeseo barojabeuseyo'' (``The wall over there is not plumb, so fix it by bracing with \emph{gakkumok}'').

\begin{table*}[t]
\centering
\small
\caption{Qualitative comparison on a sentence containing the key term \emph{gakkumok} (squared timber used as a brace). The correct counterpart is ``timber braces.''}
\label{tab:qual}
\begin{tabular}{lp{5.2cm}p{7.2cm}}
\toprule
Model & Translation & Qualitative analysis \\
\midrule
M2M100 baseline & ``If the wall does not fit, repair it.'' & Fails on the key term. Oversimplifies the context and does not convey a concrete work instruction. \\
M2M100 + LoRA & ``If the vertical wall does not fit, correct the curtain.'' & Mistranslates the term as the unrelated word ``curtain,'' exposing the risk of small-scale fine-tuning. \\
Gemma 3 4B (zero-shot) & ``Please adjust the position of the gaguk\ldots'' & Transliterates the term as ``gaguk'' and fails to convey its professional meaning; sentence structure is comparatively preserved. \\
Gemma 3 4B + RAG (proposed) & ``The vertical of the wall on that side is not aligned, so brace it with timber braces and fix it.'' & Retrieves the exact counterpart ``timber braces'' from the knowledge base and translates the instruction most precisely and clearly. \\
\bottomrule
\end{tabular}
\end{table*}

This case shows that RAG provides far more reliable and accurate translation of concrete, important site terms.
Whereas the LoRA-tuned model distorted the meaning with an inaccurate translation (``curtain''), RAG avoided that risk of fine-tuning and secured a correct translation through externally verified knowledge.
The evidence is a single illustrative example and should not be read as a measure of error rates.

\section{Limitations, Future Work, and Conclusion}
\label{sec:conclusion}

\textbf{Limitations and future work.}
The proposed sLLM+RAG system is a tool that translates passively whatever the user inputs.
To raise site safety more fundamentally it should become proactive.
As future work we propose a \emph{cognitive personal protective equipment (Cognitive PPE)}: the system is additionally trained on safety guidelines so that it supplies safety rules actively rather than only translating.
For example, given the input ``start work at height,'' the system could append to its translation an automatic warning such as ``Warning: Ensure fall protection is in place.''
Such a function would provide, in real time, preventive measures against the accident types that recur in construction disaster statistics and would turn the translation system from a mere communication tool into something like essential protective equipment for workers, a next-generation safety-management paradigm that recognizes and prevents hazards beforehand.

\textbf{Conclusion.}
With the goal of preventing accidents on construction sites, we proposed a real-time domain-specific translation system that combines a lightweight language model (sLLM) with retrieval-augmented generation (RAG) and verified it empirically.
We presented a methodology for efficiently building a construction-domain dataset with LLM-based synthetic data generation.
The proposed Gemma 3 4B + RAG model achieved the highest COMET score, the metric of semantic accuracy, among all compared models, including the multilingual translation models and the fine-tuned models, while the fine-tuned models retained higher BLEU and chrF.
For a task in which exact terminology matters, referencing externally verified knowledge in real time is thus a more efficient and maintainable solution than internalizing knowledge in the model parameters through fine-tuning.
The study offers a practical and extensible framework for communication tools in industrial settings with strong specificity and safety requirements, and we expect the proposed system to contribute to both safety and productivity on site.

\bibliographystyle{IEEEtran}
\bibliography{refs}

\appendices

\section{Key Construction Terms in the Knowledge Base}
\label{app:terms}

\cref{tab:terms} lists the 50 terms of the knowledge base.
Terms are given in Revised Romanization; the Korean forms are provided in the accompanying term list (\texttt{terms\_ko\_en.csv}).
``Meaning'' is the standard Korean counterpart rendered in English, ``English'' is the translation stored in the knowledge base, and the definitions have been translated from Korean.

\begin{table*}[t]
\centering
\scriptsize
\caption{The 50 site-slang terms used in the study.}
\label{tab:terms}
\setlength{\tabcolsep}{3pt}
\begin{tabular}{@{}r l l l p{7.6cm}@{}}
\toprule
\# & Term & Meaning & English & Definition \\
\midrule
1 & \emph{gakkumok} & squared lumber & Timber Brace & Squared timber used for structural reinforcement \\
2 & \emph{gadawakku} & formwork & Formwork & Temporary mold that shapes poured concrete or structures \\
3 & \emph{gabe} & wall & Wall & Vertical structure that divides space or bears load \\
4 & \emph{gappa} & waterproof sheet & Tarpaulin & Waterproof cover that protects materials and equipment from rain and moisture \\
5 & \emph{gombang} & manual carrying & Manual Transport & Carrying materials directly by hand or with a handcart \\
6 & \emph{darukki} & rafter & Rafter & Inclined timber member supporting a roof structure \\
7 & \emph{demodo} & assistant worker & Helper & Worker who assists skilled tradespeople on site \\
8 & \emph{dekoboko} & uneven surface & Uneven Surface & Surface that is not level and needs leveling work \\
9 & \emph{denjo} & ceiling & Ceiling & Structure or finished surface forming the top of an interior space \\
10 & \emph{dokkidasi} & terrazzo finish & Terrazzo Finish & Floor finishing method using marble or stone chips mixed in \\
11 & \emph{dobi} & hauling worker & Heavy Lifter & Worker specialized in carrying heavy equipment or large materials \\
12 & \emph{redyusya} & pipe reducer & Reducer & Reducing fitting that joins pipes of different diameters \\
13 & \emph{rube} & cubic meter & Cubic Meter (m$^3$) & Unit of volume, 1 m $\times$ 1 m $\times$ 1 m \\
14 & \emph{meji} & joint & Grout & Material that fills the gaps between bricks or tiles \\
15 & \emph{midasi} & concrete finishing & Finishing Concrete & Smoothing the surface of placed concrete \\
16 & \emph{barasi} & dismantling & Demolition & Taking apart structures, formwork, and the like \\
17 & \emph{bando} & sling belt & Sling Belt & Belt used when lifting materials with a crane \\
18 & \emph{bansaeng} & binding wire & Binding Wire & Wire used to tie and fix rebar \\
19 & \emph{borubang} & bench drill & Drill Press & Stationary drill for precision holes \\
20 & \emph{beurekka} & concrete breaker & Breaker & Equipment that breaks up concrete or asphalt \\
21 & \emph{ppade} & putty & Putty & Material that fills gaps or holes before painting or finishing \\
22 & \emph{ppalu} & pry bar & Crowbar & Lever used for demolition and nail removal \\
23 & \emph{sappodo} & formwork support & Formwork Support & Structure that fixes and supports formwork \\
24 & \emph{sasikkomi} & embedded outlet or insert & Inserted Fitting & Electrical outlet or fitting embedded in walls \\
25 & \emph{sanseunggak} & three-sun lumber & Square Timber & Lumber with a cross-section of about 3 sun (about 9 cm) \\
26 & \emph{seumi} & ink line & Ink Line & Reference line marked for drawing straight lines \\
27 & \emph{seupesya} & spacer & Spacer & Part that keeps rebar spacing constant \\
28 & \emph{seullabeu} & concrete slab & Slab & Horizontal concrete structure forming floors or roofs \\
29 & \emph{sirusi} & marking & Marking & Mark that indicates a work position or reference \\
30 & \emph{siage} & finishing & Finishing & Final finishing work that completes the appearance of a structure \\
31 & \emph{sseumi} & stacking & Stacking & Stacking materials in an orderly way \\
32 & \emph{asiba} & scaffold & Scaffold & Temporary working platform for work at height \\
33 & \emph{amrolbakseu} & waste container & Dump Container & Large steel box for collecting and hauling construction waste \\
34 & \emph{yagiri} & wall formwork & Wall Formwork & Large formwork for placing wall concrete \\
35 & \emph{yarigada} & position and height reference frame & Guideline Frame & Reference frame that sets the position and height of a building \\
36 & \emph{yarikkiri} & leave-when-done contract work & Task-based Daily Work & System in which workers leave once the assigned workload is finished \\
37 & \emph{odoriba} & stair landing & Stair Landing & Flat resting area midway on a staircase \\
38 & \emph{obikki} & three-sun lumber & Square Timber & Lumber of the same size as sanseunggak \\
39 & \emph{uma} & sawhorse & Horse Scaffold & Small movable scaffold for low work \\
40 & \emph{umemodosi} & backfilling & Backfilling & Refilling an excavation with soil or gravel \\
41 & \emph{yuka} & floor & Floor & Floor structure that people walk on \\
42 & \emph{jendai} & brick shelf & Brick Shelf & Shelf built into a wall by masonry \\
43 & \emph{tubai} & 2$\times$4-inch lumber & 2$\times$4 Timber & Standard lumber widely used in wood framing \\
44 & \emph{habakki} & baseboard & Skirting Board & Finishing material at the junction of wall and floor \\
45 & \emph{haseuri} & removal of protrusions & Chipping & Removing protrusions from a concrete surface \\
46 & \emph{hanggong madae} & large sack & Jumbo Bag & Bag for large quantities of materials or waste \\
47 & \emph{hera} & putty spreader & Putty Knife & Tool for applying putty or mortar \\
48 & \emph{hebe} & square meter & Square Meter & Unit of area (m$^2$) \\
49 & \emph{huang} & ventilation opening & Ventilation Fan & Device that circulates and exhausts indoor air \\
50 & \emph{hisasi} & eaves or canopy & Eaves, Canopy & Projecting roof structure that shields from sun and rain \\
\bottomrule
\end{tabular}
\end{table*}

\end{document}